\documentclass{article}

\usepackage{microtype}
\usepackage{graphicx}
\usepackage{subfigure}
\usepackage{booktabs} 
\usepackage{mathtools}
\usepackage[ruled,vlined,algo2e]{algorithm2e} 
\usepackage{caption}
\usepackage{epstopdf}
\usepackage{amsmath,amsfonts,amssymb,amsthm}
\usepackage{hyperref}
\usepackage{nameref}
\usepackage{graphicx}
\usepackage{amsmath,amssymb}
\usepackage{dsfont}
\usepackage{hyperref}

\usepackage{amsthm}
\usepackage{dsfont}
\usepackage{epstopdf}
\usepackage{mathtools}
\usepackage{amsmath,amsfonts,amssymb}
\usepackage{hyperref}
\usepackage{nameref}
\usepackage{xcolor}
\usepackage{paralist}
\usepackage{microtype}
\usepackage[compact]{titlesec}
\usepackage{tabularx}
\usepackage{textcomp}
\usepackage{float}
\usepackage[cal=cm]{mathalfa}
\usepackage{semantic}
\usepackage{booktabs}
\usepackage{multirow}
\usepackage{tikz}
\usepackage{wrapfig}
\usepackage{lipsum}
\usepackage{booktabs}

\usepackage[accepted]{icml2021}

\begin{document}

\twocolumn[
\icmltitle{Causal-aware Safe Policy Improvement for Task-oriented dialogue}



\centering
\begin{icmlauthorlist}
\icmlauthor{Govardana Sachithanandam Ramachandran}{}
\icmlauthor{Kazuma Hashimoto}{}
\icmlauthor{Caiming Xiong}{}
\end{icmlauthorlist}



\centering
Salesforce Research \\
\{gramachandran, k.hashimoto, cxiong\}@salesforce.com


\vskip 0.3in
]



\let\thefootnote\relax\footnotetext{This work is under review}
\footnotetext{The code available at: https://github.com/salesforce/CASPI}
\begin{abstract}
The recent success of reinforcement learning's (RL) in solving complex tasks is most often attributed to its capacity to explore and exploit an environment where it has been trained.
Sample efficiency is usually not an issue since cheap simulators are available to sample data on-policy.
On the other hand, task oriented dialogues are usually learnt from offline data collected using human demonstrations.
Collecting diverse demonstrations and annotating them is expensive.
Unfortunately, use of RL methods trained on off-policy data are prone to issues of bias and  generalization, which are further exacerbated by stochasticity in human response and non-markovian belief state of a dialogue management system.
To this end, we propose a batch RL framework for task oriented dialogue policy learning: causal aware safe policy improvement (CASPI). This method gives guarantees on dialogue policy's performance and also learns to shape rewards according to intentions behind human responses, rather than just mimicking demonstration data; this couple with batch-RL helps overall with sample efficiency of the framework. We demonstrate the effectiveness of this framework on a dialogue-context-to-text Generation and end-to-end dialogue task of the Multiwoz2.0 dataset. The proposed method outperforms the current state of the art on these metrics, in both case. In the end-to-end case, our method trained only on 10\% of the data was able to out perform current state in three out of four evaluation metrics.
\end{abstract}
\section{Introduction}

\begin{figure}
    \scalebox{.3}{
    \centering
    \includegraphics{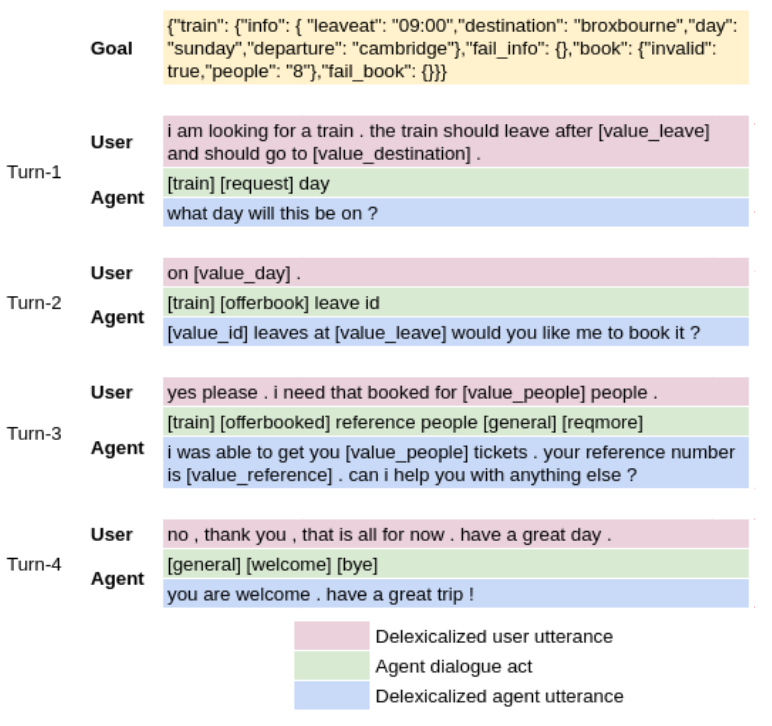}}
    \caption{A typical Task oriented dialogue conversation in MultiWoz2.0 dataset}
    \label{fig:todEx}
\end{figure} 
Offline task-oriented dialogue systems involves solving disparate tasks of belief states tracking, dialogue policy management, and response generation. In this work we strive to improve the performance of dialogue policy management. The need for sample efficiency sample efficiency is key for learning offline Task oriented dialogue system as the access to data are finite and expensive. Recent advancements in Off-policy based reinforcement learning (Batch-RL) methods that uses historical annotated data as against a simulator has proven to be sample efficient and helps in safe policy improvement for generalizable policies. The effective use of these techniques are hindered by the nature of dialogue policy learning. For example, Off-policy based learning many times requires an estimation of behaviour policy for a given state of Markov Decision Process (MDP). In real life, belief-state does not capture the true state of the MDP latent state such as prosody, among others induce stochasticity in the agents response at each turn. Then there is the issue of loss of semantic information from dialogue act to generated natural language text. This is demonstrate by Fig: \ref{fig:todEx}. Use of mere policy imitation for dialogue-act falls short of reasoning on the outcome, rather focuses on each constituent of composite action equally. This is demonstrated in Fig:\ref{fig:todEx}. Turns\#3 and \#2 are rich in semantic information and Turn\#3 key to the transaction of the booking process, while Turn\#4 though of least use in the success of the conversation gets equal weight as other semantically rich term, worse the appear more often than specifics like Turn\#2 and \#3 there by clogging the gradient budget. These importance are lost in imitation policy learning.   
\begin{figure*}
    \scalebox{.35}{
    \centering
    \includegraphics{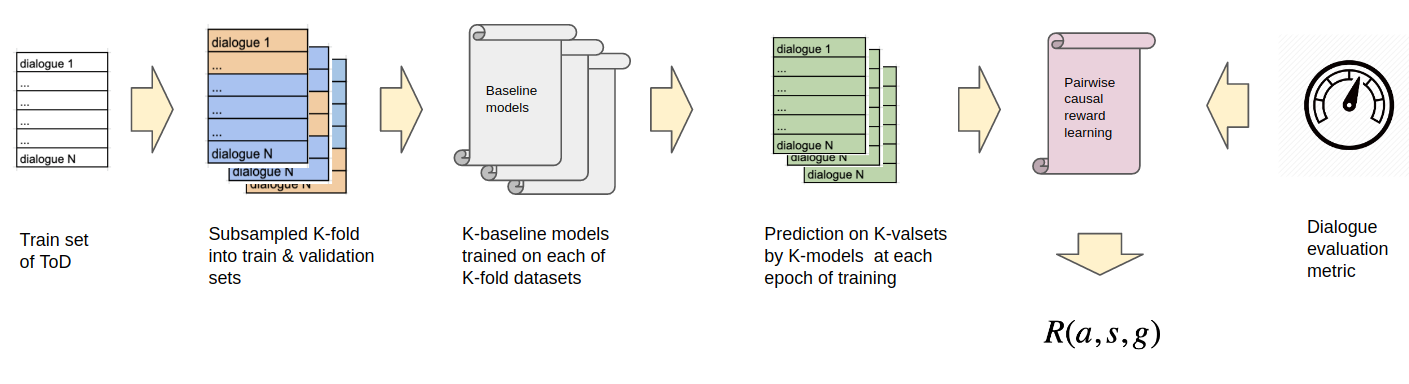}}
    \caption{Process flow of pairwise causal reward learning}
    \label{fig:pairwise_flow}
\end{figure*} 

The main contribution of this work are, we introduce safe policy improvement in batch reinforcement setting for dialogue policy learning with guarantees for performance. We introduce pairwise causal reward learning to shape reward that reason the intention of human utterance instead of mimic the demonstration. By use of these two off-policy methods we demonstrate sample efficiency. 
\section{Related Works}

With the release of multi-domain, multi-turn MultiWoz2.0 dataset~\cite{multiwoz}, there has been flurry of recent works, of which \cite{damd} uses data augmentation.
\citet{schemaGuided} and \citet{simpletod} frame  dialogue policy learning as language modeling task. Among the works that uses reinforcement learning. \citet{slrl} uses supervised learning to bootstrap followed by RL fine tuning, whereas \cite{larl} uses policy gradient on latent action space as against handcrafted ones. To our best of knowledge \cite{wayOffPolicy} and \cite{hdno} the only other work that uses Batch-RL for dialogue policy learning. Recently there's has been proliferation in use of large pretrained language model based systems like \cite{simpletod} \cite{mintl} \cite{HDSA} etc.

The line of inverse RL used in this work can be traced back to \citet{maxEntropyIRL}, proposes roll-outs from expert demonstration should have rewards exponentially higher than any other arbitrary roll-outs. This method requires a normalizing constant that integrates across rollouts, which is challenging. \citet{humanprefernce} and \citet{rankingIRL} propose to do relative comparison of two roll-outs there by eliminating the need for normalization constant and they demonstrate in online setting. 
\section{Method}
\begin{figure*}
    \centering
     \includegraphics[width=.8\textwidth\relax]{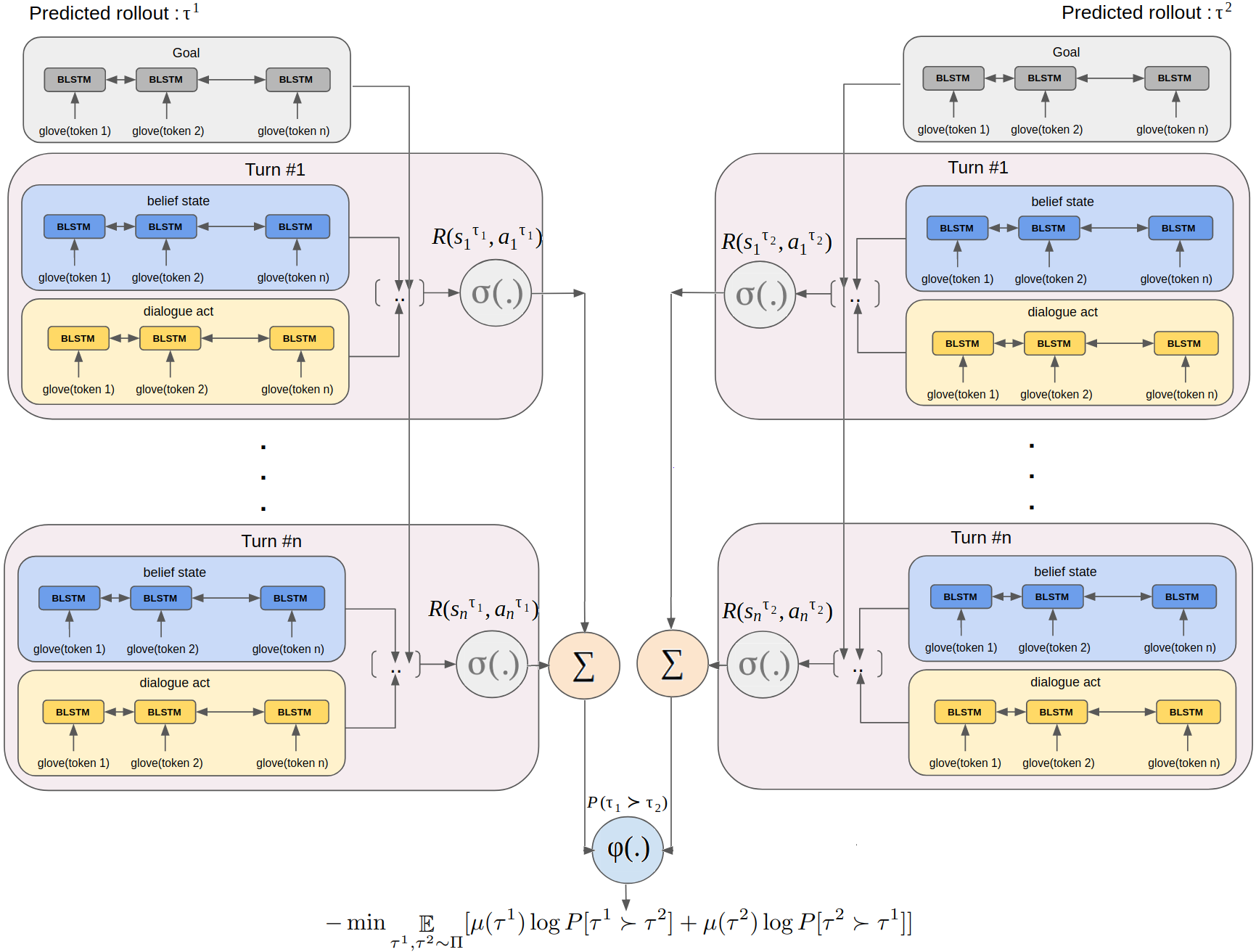}
    \caption{Pairwise causal reward learning network architecture}
    \label{fig:pairwise_model}
\end{figure*} 
\begin{figure}
    \scalebox{.37}{
    \centering
    \includegraphics{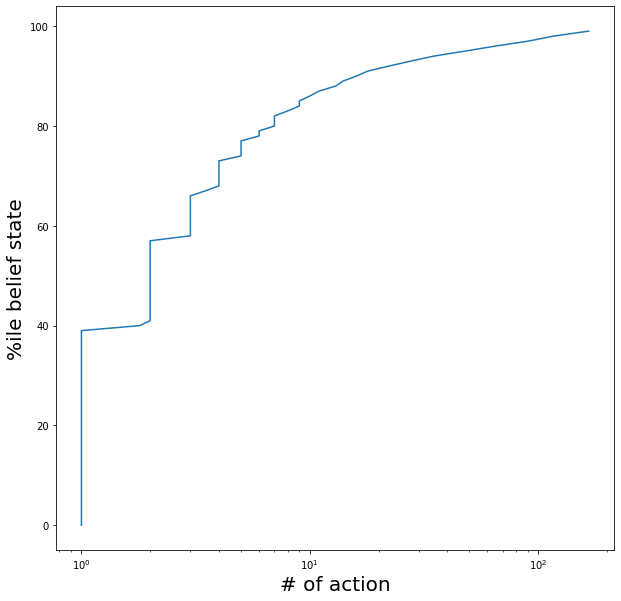}}
    \caption{Shows stochacity of dialogue act against belief state in the MultiWoz2.0 dataset}
    \label{fig:stoch_belief}
\end{figure} 
\subsection{Preliminaries}

We model task-oriented dialogue as a Markov decision process (MDP) \cite{sutton2018reinforcement} with set of states $S$ and actions $A$. The agent at time step $t$ with state $s_t$ performs a composite action $a_t$ as per a target policy $\pi_e(a_t|s_t)$ on the environment with transition probabilities to next state $P(s_{t+1}|s_t,a_t)$, a latent reward function, $R(s_t, a_t)$ with discount factor $\gamma \in [0,1]$. Then the objective is to optimize for the target policy $\pi_e$, that maximizes the discounted sum of future reward on the MDP, given the state-action value function $Q^{\pi_e}(a_t,s_t) = \mathop{\mathbb{E}}_{a_t \sim \pi_e, s_t \sim P}[\sum_{t'=t}^{T}\gamma^{t-t'}R(s_{t'},a_{t'})]$.

In offline Batch-RL. The agent does not get to interact with the environment, instead we are provided with offline data $D$ logged by human agents performing actions based on a latent stochastic behaviour policy $\pi_b$, where $\tau^i \in D$ is a rollout of a dialogue, composing of $\tau^i = ((o_0^i,a_0^i),...,(o_{T-1}^i,a_{T-1}^i))$. Here the $o_t$ is the observation at turn $t$, composing of $o_t=(b_t,u_t^u,u_{t-1}^a)$, where $b_t$ is the belief state of the agent at turn $t$, $u_t^u$ and $u_{t-1}^a$ are the user and agent utterance at time $t$ and $t-1$ respectively.

\subsection{Safe policy improvement}
Batch-RL entails training a policy on rollout generated by the latent behaviour policy. Directly optimizing on the rollouts generated by another policy, leads to large bias in the value function estimation, poor generalization characteristic, and sample inefficiency \cite{batchRLsampleeff}. Safe policy improvement ensures the new policy performance is bounded compared to the old policy, as in this case the behaviour policy. This is given by:
\begin{equation*}
Pr(V^{\pi_e} \geq V^{\pi_b} - \zeta) \geq 1-\delta,
\end{equation*}
where $V^{\pi_e}$ and $V^{\pi_b}$ are value functions of the target and behaviour policy respectively. Here $1-\delta$ and $\zeta$ are the high probability and approximation meta-parameters respectively. \citep{trpo} provide such update mechanism, \eqref{trpoeqn}, whose errors are bounded as long as the constraints of \eqref{trpoeqn} are met, where $D_{KL}(.||.)$ is the KL divergence and $\eta$ is a hyper-parameter. 

\begin{equation}\label{trpoeqn}
\begin{split}
L_{sto}(\theta) = \min - \mathop{\mathbb{E}}_{s \sim P^{\pi_{b}}, a \sim \pi_b} \left[\frac{\pi_e(a|{b_t;\theta})}{\pi_b(a|{b_t})}Q^{\pi_{b}}(b_t,a_t)\right] \\
s.t.  \mathop{\mathbb{E}}_{s \sim P^{\pi_b}} [D_{KL}(\pi_b(.|{b_t})||\pi_e(.|{b_t}))] \leq \eta 
\end{split}
\end{equation}

Use of this update rule requires access to the behavior policy $\pi_b(a_t|s_t)$ which is intractable to estimate and the learnt ones might have bias. Using them to perform bias correction like Important Sampling \cite{importantsampling} might lead to worse policy. Instead we estimate the behaviour policy conditioned on the belief state $b_t$ as against $s_t$ in \eqref{trpoeqn}, which is result in a stochastic behavior policy. The belief state $b_t$ is part of the observation $o_t$ at turn $t$. The actions are stochastic in nature given just the belief state, this demonstrated by Fig:\ref{fig:stoch_belief}. We purport that on availability of more evidence of the observation $o_t$, (beside $b_t$) the mode of the policy collapse to a near deterministic action.  To  factor this into the policy learning, we have an additional loss: 
\begin{equation}
L_{det}(\theta)=\min - \mathop{\mathbb{E}}_{(o_t,a_t) \sim D}[G(\tau,t)\log\pi_e(a_t|o_t)],
\end{equation}
where $G(\tau^1, t)=\sum_{t' = t}^{T} \gamma^{t'-t} R(s_{t'}^1,a_{t'}^1,g^1)$ is the discounted sum of future reward for a rollout $\tau^1$ with goal $g^1$.Hence policy optimization loss function is given by:
\begin{equation}\label{totalLoss}
L(\theta) = L_{sto}(\theta) + L_{det}(\theta)
\end{equation}
We achieve this by doing two forward passes on the policy network, first with only the belief state as the input and another pass with all the observation information to policy network to get the action distribution. The first pass captures the stochasticity of the policy conditioned only on the belief state, $b_t$ and the second pass collapse the mode given other latent information of the state, such as $u_{t}^u$ and $u_{t}^a$.  

\subsection{Pairwise causal reward learning}
\begin{algorithm}[tb]
   \caption{CASPI}
   \label{alg:caspiAlgo}
\begin{algorithmic}
    \STATE \textbf{Input}: Dialogue dataset $D$ and evaluation metric $M$  
    \STATE Sub-sample K-folds of train and val set $(D_T,D_V) \sim D$ 
    \FOR{ $\forall (D_T,D_V)$ }
        \STATE Learn ToD in supervised setting by optimizing objective:
        \STATE $ -\min \mathop{\mathbb{E}}_{a,s \sim D_T} \log(\pi_m(\hat{a}|s))$
        \FOR{ $\forall$ epoch}
            \STATE Predict on the valset $D_V$ and add it to the dataset, $D_P$ for pairwise causal learning
            \STATE $D_P=D_P\cup\tau|\tau \sim \pi_m$
        \ENDFOR
    \ENDFOR
    \REPEAT 
        \STATE Sample pair of rollouts $(\tau^1,\tau^2) \sim D_P$
        \STATE Learn for $R(.)$ network by optimizing for objective Eqn: \ref{pairwiseobj}
    \UNTIL{Convergence using data $D_P$}
    \REPEAT
        \STATE Optimize for policy $\pi_e$ using objective \ref{totalLoss}
    \UNTIL{Convergence using data $D$}
\end{algorithmic}
\end{algorithm}
\label{pairwiseMethod}
\begin{table*}
\centering
\begin{tabular}{llllllll}
\hline
\textbf{Model} & \textbf{Belief} & \multicolumn{2}{c}{\textbf{System Action}} & \textbf{Inform} & \textbf{Success} & \textbf{BLEU} & \textbf{Combined} \\
& \textbf{State  Type} & \textbf{Type} & \textbf{Form} & \textbf{(\%)} & \textbf{(\%)} & & \textbf{Score} \\
\hline
HDSA \citep{HDSA} & \small\verb|Oracle| & \small\verb|generated| & \small\verb|graph| & \small\verb|82.9| & \small\verb|68.9| & \small\textbf{23.6} & \small\verb|99.50| \\
\hline
LaRL\citep{larl} & \small\verb|Oracle| & \small\verb|generated| & \small\verb|graph| & \small\verb|82.8| & \small\verb|79.2| & \small\verb|12.8| & \small\verb|93.80| \\
\hline
DAMD\citep{damd} & \small\verb|Oracle| & \small\verb|generated| & \small\verb|span| & \small\verb|89.2| & \small\verb|77.9| & \small\verb|18.6| & \small\verb|102.15|\\
\hline
SOLONIST\citep{damd} & \small\verb|Oracle| & \small\verb|generated| & \small\verb|span| & \small\verb|89.6| & \small\verb|79.3| & \small\verb|18.3| & \small\verb|102.75|\\
\hline
MarCo\citep{damd} & \small\verb|Oracle| & \small\verb|generated| & \small\verb|span| & \small\verb|92.3| & \small\verb|78.6| & \small\verb|20.02| & \small\verb|105.47|\\
\hline
HDNO\citep{damd} & \small\verb|Oracle| & \small\verb|generated| & \small\verb|span| & \small\verb|96.4| & \small\verb|84.7| & \small\verb|18.85| & \small\verb|109.40|\\
\hline
CASPI(DAMD), $M_{soft}(act)$ & \small\verb|Oracle| & \small\verb|generated| & \small\verb|span| & \small \textbf{96.8} & \small\textbf{87.3} & \small\verb|19.10| & \small\textbf{111.15}\\
\hline

\end{tabular}
\begin{*}
\caption{\label{context2resp}
Comparison of results for dialogue-context-to-text generation task  of Multiwoz2.0. The use of ground truth or generated results are denoted as Oracle and generated respectively.
}
\end{*}
\end{table*}
The policy optimization objective introduced in the previous section requires access to 
per time-step reward $R(s_t,a_t,g))$. To this end, we provide a mechanism to learn a reward that is causally reasoned on the intention of the human demonstrator.
Usually dialogue policy learning is accompanied by metrics $M$, to evaluate the performance of the learnt policy. Though these metrics could serve as a proxy for a reward function, using them directly is challenging. These metric functions usually returns a score for the entire dialogue. Given the complex state-action space of the dialogue management system, these dialogue level feedback are under-specified for rewarding an action performed at each turn. 

To address this under-specified feedback, we adapt the preference learning introduced by  \cite{humanprefernce} from an online to an offline setting. We parametrize reward for every timestep $t$, as $R(s_t,a_t, g)$. Given a pair of rollouts $\tau^1,\tau^2 \in D$ with actions for each state in the rollouts sampled from the different learnt policies $\pi^1_e$ and $\pi^2_e$ respectively. Let $P[\tau^1 \succ \tau^2]$ be the probabilistic measure that captures the preference for policy $\pi^1_e$ over policy $\pi^2_e$. This preference is true when the sum of rewards of each dialogue of the two rollouts is such that. $\sum_{t = 0}^{T} R(s_t,a_t|(s_t,a_t)\in\tau^1)  >\sum_{t = 0}^{T} R(s_t,a_t,g|(s_T,a_t)\in\tau^2)$. We henceforth we refer $\sum_{t = 0}^{T} R(s_t,a_t,g|(s_T,a_t)\in\tau)$ as $R(\tau)$ Then the preferential probability can be represented by:

\begin{equation*}
P[\tau^1 \succ \tau^2] = \frac{\phi(R(\tau^1))}{\phi(R(\tau^1))+\phi(R(\tau^2))}
\end{equation*}

Here $\phi(.)$ could either be $exp(.)$ or identity $\mathds{1}(.)$. In our experiments later works best. We optimize for reward, $R(s_t,a_t,g)$ by minimizing binary cross-entropy loss between the preference probability and the normalized metrics score, $\mu(\tau)$ between a pair of rollout.

\begin{equation}
\begin{split}
L(\theta) = \min -\mathop{\mathbb{E}}_{\tau^1,\tau^2 \sim \Pi} [\mu(\tau^1) \log P[\tau^1 \succ \tau^2] \\ + \mu(\tau^2) \log P[\tau^2 \succ \tau^1]] 
\end{split}
\label{pairwiseobj}
\end{equation}

where,
\begin{equation}
\mu(\tau^1)=\frac{M(\tau^1)}{M(\tau^1)+M(\tau^2)}
\label{normalizedScore}
\end{equation}

Learning policy in a sparse reward MDP is a hard problem \cite{explorego}. In online learning, the agents can interact and explore the environment. The agents have the liberty to sample arbitrarily large numbers of rollouts from the environment and it may still fail \cite{explorego} to learn effective policy in sparse reward MDP with large state-action space, as the chance of encountering non-zero reward grows exponentially smaller with the episode length.

This is exacerbated in offline settings as we are forced to learn optimal policy with finite data. Some successes are seen with guided exploration \cite{hardVideo} \cite{sparseRewardGuided1} \cite{sparseRewardGuided2}, where expert demonstrations are used to guide the exploration. This strategy improves the chance of encountering the sparse reward as it restricts the state-action space to regions where non-zero reward exists.

We observe that the dialogue roll-outs are generated by expert latent policy. The data (dialogue rollouts) are distributed as per the optimal latent policy and transition probability. We propose that predictions made by a policy while in the process of learning to maximize the likelihood of the data is a good curriculum for exploring the state-action space for pairwise reward learning. This is the key insight of this work.

We formalize this insight into a method depicted in Fig:\ref{fig:pairwise_flow} and Algo:\ref{alg:caspiAlgo}. The (train) dataset is subsampled into $K$-fold train \& val sets. $K$-baseline models are trained to fit the data distribution generated by experts using cross entropy loss. During the process of fitting the data distribution, the still learning K-policies are used to predict on their corresponding K-fold valset  at every epoch of the training. Each of the dialogue is scored by the chosen dialogue level metric. On convergence of the supervised learning process. Pairs of dialogue predictions generated by the above process, along with their corresponding metric score are used to train for preferential optimization objective Eqn.\ref{pairwiseobj}, which in-turn learns fine grained reward $R(a,s,g;\theta)$. The use of K-fold subsampling and K-baseline models helps generate stochaticity in the samples generated. It also helps in effectively using the data and make the method sample efficient. 
\begin{table*}
\centering
\begin{tabular}{lccccc}
\hline
\textbf{Model} & \textbf{Pre-trained model} & \textbf{Inform \%} & \textbf{Success \%} & \textbf{BLEU} & \textbf{Combined Score} \\
\hline
DAMD & \small\verb|No| &  \small\verb|72.79| & \small\verb|60.45| & \small\verb|16.93| & \small\verb|83.55|\\
\hline
DAMD + multi-action & \small\verb|No| & \small\verb|76.33| & \small\verb|64.35| & \small\verb|17.96| & \small\verb|88.30|\\
\hline
SimpleTOD & \small\verb|Yes| &  \small\verb|84.4| & \small\verb|70.10| & \small\verb|15.01| & \small\verb|92.26|\\
\hline
SOLOIST & \small\verb|Yes|  & \small\verb|85.5| & \small\verb|72.90| & \small\verb|16.54| & \small\verb|95.74|\\
\hline
MinTL-BART & \small\verb|Yes|  & \small\verb|84.88| & \small\verb|74.91| & \small\verb|17.89| & \small\verb|97.79|\\
\hline
CASPI(DAMD), $M_{soft}(act)$ & \small\verb|No|  & \small\verb|89.1| & \small\verb|76.1| & \small\textbf{18.08} & \small\verb|100.68|\\
\hline
CASPI(MinTL), $M_{soft}(act)$ & \small\verb|Yes|  & \small\textbf{94.59} & \small\textbf{85.59} & \small\verb|17.96| & \small\textbf{108.05}\\
\hline
CASPI(MinTL), $M_{hard}(act)$ & \small\verb|Yes|  & \small\verb|93.79| & \small\verb|84.88| & \small\verb|17.47| & \small\verb|106.81|\\
\hline
\end{tabular}
\begin{*}
\caption{\label{end2endResults}
Comparison of results for end-to-end task  of Multiwoz2.0. 
}
\end{*}
\end{table*}

\subsection{Sample weights for policy optimization}
\begin{equation}
\theta := \theta -  R_{caspi}(s,a) \nabla \pi_{blackbox}(a|s;\theta)  
\label{sampleWt}
\end{equation}
The learnt reward is akin to sample weights for each instance of the data, that helps to redistribute the gradient update budget among the samples based of their contribution to the the overall success of the Task oriented Dialogue system. To this end, we propose that learnt reward could be used as sample weight to any existing ToD dialogue system to reap the benefit of sample efficiency it brings. We demonstrate this by adopting two exiting ToD with the learnt reward, more about this in the next section \ref{model}  
\section{Experimental Settings}

\begin{table*}
\centering
\begin{tabular}{llllllllll}
\hline
\textbf{Model}  & \multicolumn{3}{c}{\textbf{5\%}}  & \multicolumn{3}{c}{\textbf{10\%}} & \multicolumn{3}{c}{\textbf{20\%}}  \\
& \textbf{Inform} & \textbf{Success} & \textbf{BLEU} & \textbf{Inform} & \textbf{Success} & \textbf{BLEU} & \textbf{Inform} & \textbf{Success} & \textbf{BLEU} \\
\hline
MD-Sequicity  & \small\verb|49.40| & \small\verb|19.70| & \small\verb|10.30| & \small\verb|58.10| & \small\verb|34.70| & \small\verb|11.40| & \small\verb|64.40| & \small\verb|42.10| & \small\verb|13.00| \\
\hline
DAMD  & \small\verb|56.60| & \small\verb|24.50| & \small\verb|10.60| & \small\verb|62.00| & \small\verb|39.40| & \small\verb|14.50| & \small\verb|68.30| & \small\verb|42.90| & \small\verb|11.80| \\
\hline
MinTL   & \small\verb|75.48| & \small\verb|60.96| & \small\textbf{13.98} & \small\verb|78.08| & \small\verb|66.87| & \small\textbf{15.46} & \small\verb|82.48| & \small\verb|68.57| & \small\verb|13.00| \\
\hline
CASPI(MinTL), $M_{soft}(resp)$  & \small\verb|87.69| & \small\textbf{71.17} & \small\verb|13.51| & \small\verb|82.08| & \small\verb|72.27| & \small\verb|14.10| & \small\verb|89.39| & \small\verb|78.58| & \small\textbf{15.16} \\
\hline
CASPI(MinTL), $M_{hard}(resp)$  & \small\textbf{89.69} & \small\verb|69.47| & \small\verb|13.33| & \small\textbf{92.59} & \small\textbf{78.58} & \small\verb|14.48| & \small\textbf{94.19} & \small\textbf{83.28} & \small\verb|13.65| \\
\end{tabular}
\begin{*}
\caption{\label{sampleEff1}
Comparison of results for end-to-end  of Multiwoz2.0. in low resource setting
}
\end{*}
\end{table*}

\subsection{Model}
\label{model}
\subsubsection{CASPI(.)}
We believe our pairwise casual reward learning and associated sample improvement is independent of model architecture used for learning Task oriented Dialogue systems. As argued in the previous section our approach could be used as sample weights for any existing methods. To this end we choose two TOD methods that are at the extremes of model architecture spectrum 1) One uses a light weight custom model and 2) Other uses a large standard pre-trained  out-of-the box universal language model. We demonstrate the ease of integrating of CASPI with these methods, and demonstrate the improvement in performance and sample efficiency.

\subsubsection{CASPI(DAMD)}

In this setting , we use the neural model proposed by \cite{damd} without their key contribution of data augmentation as the baseline for our experiments. DAMD is composed of three $seq2seq$ generative model using GRUs. The three $seq2seq$ models are one each for belief state, dialogue act and response generation modules. An attention layers is then used to attend the outputs of the $seq2seq$ models with the context vector of previous turn for copy over mechanism. The outputs are then used as representation for predicting series of tokens for their respective modules. For more details on the model architecture and parameter setting refer \cite{damd}. In this setting we use both stochastic, $L_{sto}$ and deterministic, $L_{det}$ loss functions on dialogue act. For DST and response generation, we retain the cross entropy loss as is from DAMD\cite{damd}.

\subsubsection{CASPI(MinTL)}

On the other extreme of model complexity, we use the Task oriented Dialogue model, MinTL\cite{mintl}. MinTL uses a large pretrained language model BART\cite{bart}. BART use as a standard encoder decoder transformer architecture  with a bidirectional encoder and an auto-regressive decoder. It is pre-trained on the task of denoising corrupt documents. BART is trained using cross-entropy loss between the decoder output and the original document. For more  details of the model architecture and parameter setting, we suggest referring to \cite{mintl} \cite{bart}.

MinTL doesn't explicitly predict dialogue act. Hence we only use the deterministic loss, $L_{det}$ directly on the generated response and for DST we retain the loss as is from MintTL \cite{mintl}.

\subsubsection{Pairwise Causal Learning Network}

Fig \ref{fig:pairwise_flow} describes the process flow of pairwise casual reward learning. We chose DAMD \cite{damd} model for it's light weight to train $K$ baseline models and in the process of training, generate rollouts for pairwise causal reward learning. In all our experiments, we use $K=10$.

Fig:\ref{fig:pairwise_model} illustrates the pairwise casual reward learning network. We use three single bi-LSTM layers, one each to encode goal, belief state and dialogue act or response sequences at each dialogue turn on each of the sampled roll-outs pairs, $\tau^1$ and $\tau^2$. The three encoded representations are concatenate and are fed through couple of feed-forward layers before making a bounded reward prediction $R(s_t, a_t, g)$ for each turn using a sigmoid function. The per turn rewards are summed to form a global reward $R(\tau)$ for the roll-out $\tau$. Using a pair of dialogue rewards $R(\tau^1)$ and $R(\tau^2)$, we compute the probabilistic preference between the roll-outs $P[\tau^1 \succ \tau^2]$ either by standard normalization or a softmax function. The output of this optimized using  crossentopy loss described in Eqn:\ref{pairwiseobj}

\subsection{Dataset}
To evaluate our proposed method on Multi-domain Wizard-of-Oz (MultiWoz) \cite{multiwoz} dataset. It is a large scale multidomain, task oriented dataset generated by human-to-human conversation , where one participant plays the role of a user while the other plays the agent.The conversations are between a tourist and a clerk at an information center. The conversations span across 7 domains including attraction, hospital, hotel, police, restaurant, taxi and train. Each dialogue is generated by users with a defined goal which may cover 1-5 domains with a maximum of 13 turns in a conversation. The dataset has 10438 dialogues split into 8438 dialogues for training set and 1000 dialogues each for validation and test set.
\subsection{Prepossessing}
We represent DB results as one-hot vectors as proposed by \cite{dbonehot}. To reduce surface-level variability in the responses, we use domain-adaptive delexicalization preprocessing proposed in \cite{delex}. As proposed in \cite{damd}, We generate delexicalized responses with placeholders for specific values which can be filled with information in DST and database.

\subsection{Metrics}

\subsubsection{Evaluation}
Since the focus of this work is sample efficiency of dialogue policy learning, we use the context-to-response generation task of Multiwoz2.0~\cite{multiwoz} and use their evaluation metrics to measure the quality of the response as primary objective and for completeness we also evaluate performance of our method on end-to-end dialogue modeling task. Both of these setting uses three evaluations metrics. These include: 1) inform rate - measures the fraction of dialogue, the system has provided the correct entity, 2) success rate - fraction of dialogues, the system has answered all the requested information and 3) BLEU \cite{bleu} - measures the fluency of the generated response. We also report the combined score $(Inform + Success) \times 0.5 + BLEU$ proposed by \citet{slrl}. All the numbers of CASPI reported in this work are median of 5 runs with different seeds. 

\subsubsection{Training}
For the metric $M$ used in pairwise causal reward learning , we use the following:
\begin{equation}
M := Inform + Success + \lambda \times BLEU
\label{metricTrain}
\end{equation}
This is very similar to combined score used in evaluation and both are equivalent when $\lambda = 2$. We introduced hyperparamter $\lambda$ to normalize the achievable scale of $BLEU$. We observe that success rate, if used as is, will result in non-markovian and stochastic per turn reward function, since the reward of current state will depend on the performance of future states. Hence, we also use a soft version of the metric $M_{soft}$, where the success rate measures a fraction of requested information provided in a dialogue. We refer the original metric that uses the discrete variant of success rate as $M_{hard}$. The choice of action in reward function $R(s_t,a_t,g)$ can either be dialogue act or generate response, we refer corresponding variants of metrics as $M(act)$ and $M(resp)$. To demonstrate the versatility of the method to adapt to different metrics, we use all the discussed variants of the metric.

\subsection{Baselines}
DAMD: Introduced by \cite{damd}is a domain-aware multi-decoder network. The method also exploits stochastic nature of the dialogue act by using a data-augmentation technique called the multi-action data augmentation. DAMD with data augmentation is denoted here as DAMD + multiaction.

HDSA by \cite{HDSA} proposes to use hierarchical graph representation for dialogue act. It uses a pre-trained 12-layer BERT model (Devlin et al., 2019) to represent dialogue act. The predicted dialogue act is transformed to the hierarchical graph structure using disentangled self-attention model, a 3-layer self-attention model (Vaswani et al., 2017)

SOLOIST \cite{soloist} and SimpleTOD \cite{simpletod} uses pretrained GPT-2-based methods. These method are trained on turn-level data without generated belief state and system act in dialog history.

MinTL-BART \cite{mintl}, introduced Levenshtein belief spans framework that predicts only the incremental change in dialogue state per turn. It leverages the pretrained T5 and BART \cite{bart} as backbone for model architecture.

HDNO proposed by \cite{hdno} is a dialogue policy learning method to solve context-to-response generation task of Multiwoz2.0 \cite{multiwoz}. It  exploits the hierarchical nature of dialogue act and response generation task by proposing an option based framework of Hierarchical RL and variational model to learn a latent dialogue act that corresponds to natural language response. Unlike our method, HDNO though highlights the risk of sparsity of metric function such as success rate as reward function, resorts to shaping a proxy reward function. Use markov language model as a proxy reward function. The language model is learnt independent of the metric function. Our method refrains from reward shaping and is independent of the nature of any underspecified metric function. Since we learn fine grained turn specific credit assignment, our solution can adapt to other metric function as long as the pairwise reward network is rich enough to factorize them.

\section{Result}
\begin{table}
\centering
\begin{tabular}{llll}
\hline
\textbf{Train data} & \textbf{Inform \%} & \textbf{Success \%} & \textbf{BLEU}  \\
\hline
100\% & 96.8 & 87.3 & 19.1 \\
75\% & 94.2 & 81.4 & 19.2 \\
50\% & 91.2 & 76.6 & 17.7 \\
25\% & 91.5 & 68.3 & 15 \\
\hline
\end{tabular}
\caption{\label{sampleEff2}
Sample efficiency study of CASPI(DAMD) on context-to-response generation task of MultiWoz2.0}
\end{table}
We first compare our method against the current state of the art methods on the context-to-response generation task defined by MultiWoz2.0, \cite{multiwoz}. The results are tabulated at Table:\ref{context2resp}. We use CASPI adaptation of DAMD, CASPI(DAMD) for this task. CASPI(DAMD) performs better than other methods on three of the four performance criteria i.e success rate, inform rate and combined score. HDSA \cite{HDSA} has better BLEU score. This rich expressiveness of natural language by HDSA, stems from the use of large 12-layers BERT \cite{bert} model.

Secondly, we compare both adaptation of our methods CASPI(DAMD) and CASPI(MinTL) on the end-to-end dialogue tasks defined by MultiWoz2.0 \cite{multiwoz}. The results are tabulated at Table:\ref{end2endResults}. CASPI(DAMD) with it's light weight model architecture with no pretraining on any external corpus, was able to out perform all other previous method in all evaluation criteria. This goes to show using CASPI to shepard the gradient update process as sample weights for each dialogue turn leads to a model that's well aligned with true objective of the task. CASPI(MinTL) with its robust pretrained model out performs CASPI(DAMD) by a large margin. This goes to show the ease of adaptation of existing methods with CASPI.  

\subsection{Sample Efficiency}
\begin{figure}
    \scalebox{.3}{
    \centering
    \includegraphics{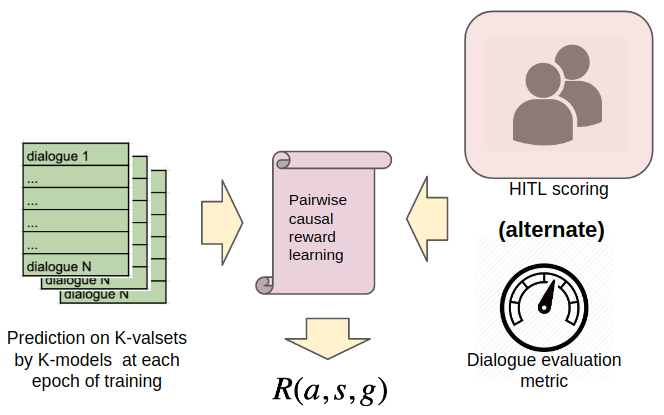}}
    \caption{Mixed Human-in-the-loop and automatic evaluation metric scores for pairwise causal reward learning}
    \label{fig:HITLLearning}
\end{figure}   
Inverse reinforcement learning, coupled with off-policy policy learning and evaluation are proven to be sample efficient \cite{batchRLsampleeff} . We argue CASPI is competitive with other sample efficiency techniques, such as data augmentation and transfer learning as performed by \cite{damd} and \cite{mintl} respectively. To demonstrate the hypothesis, we test our method against baseline in a low sample complexity regime. For experimental setup, we adopt the low resource testing strategy from \cite{mintl}. We train our model on 5\%, 10\%, and 20\% of the training data and compare with other baselines on end-to-end dialogue and context-to-response generation tasks, Table \ref{sampleEff1} and \ref{sampleEff2} list the results. In end-to-end task, CASPI(MinTL) trained only on 10\% of data was able to out perform previous state of the art method, MinTL trained on 100\% data on two of the three performance metrics. On the context-to-response generation task, CASPI(DAMD) trained on 75\% of the data was able to match 100\% data performance of HDNO. This goes to show that having the right reward function to guide the budget of the gradient update process to reach the true objective is important in extremely low resource setting.
\subsection{Human Evaluation}
\begin{figure}
    \centering
    \begin{subfigure}
        \centering
        \includegraphics[width=0.5\textwidth]{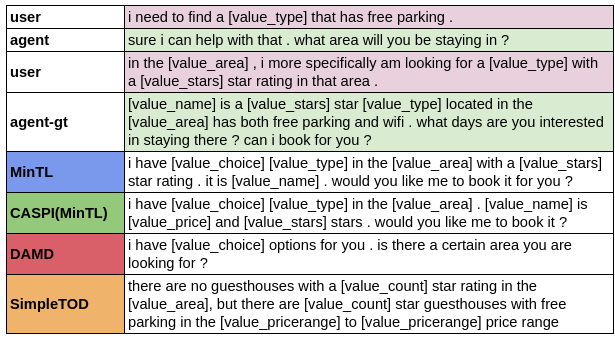}
        \caption{Example of generated responses by different ToD models}
        \label{fig:HumanEvalIllustration}
    \end{subfigure} 
    \begin{subfigure}
        \centering
        \includegraphics[width=0.5\textwidth]{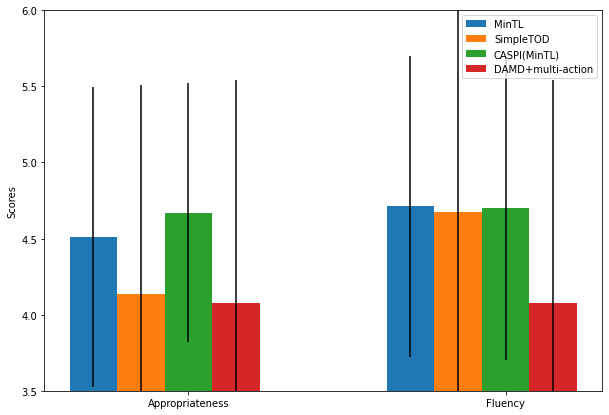}
        \caption{Human evaluation on criterias:Appropriateness and Fluency }
        \label{fig:HumanEvalResults}
     \end{subfigure}
     \begin{subfigure}
        \centering
        \includegraphics[width=0.5\textwidth]{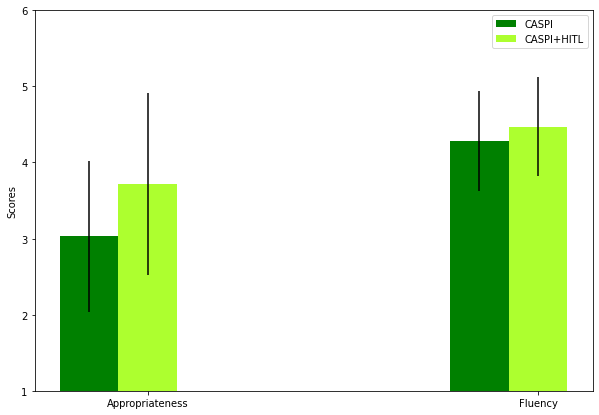}
        \caption{Human evaluation of Human in the loop training of CASPI(MinTL) on 5\% of Multiwoz2.0 dataset}
        \label{fig:HITLHumanEvaluation}
    \end{subfigure}     
\end{figure} 
Automatic evaluation metrics have their own biases. True objective of ToD is human experience while interacting with the dialogue systems, which automatic evaluation metrics might fall short to capture. To this end we conduct human evaluation on the quality of the generated response. We define quality by the following criterias:

1) Appropriateness:  Are the generated responses appropriate for the given context in the dialogue turn? 

2) Fluency: Are the generated responses coherent and comprehensible? 

A dialogue turn in the test set is randomly picked. The human evaluators were shown context leading up to the turn. The predictions for the turn by different models were anonymized and displayed to the evaluators. This is illustrated in Fig:\ref{fig:HumanEvalIllustration}. The human evaluators were asked to give a score between 1 and 5 for appropriateness and fluency, with score of 5 being best and 1 being the worst. 100 randomly selected dialogue turns were presented to 10 participants .We report the mean and variance of the score. We compare our model performance against MinTL \cite{mintl}, SimpleTOD \cite{simpletod} and DAMD \cite{damd}. Fig:\ref{fig:HumanEvalResults} shows the results of the evaluation. CASPI(MinTL) outperforms all other models in appropriateness score. While fluency score of CASPI(MinTL), MinTL and SimpleTOD are comparable to each other.

\subsection{Human in the loop training}
\label{hitlSec}
In the previous section we argue automatic dialogue evaluation metrics are biased and doesn't truly reflect the human objective but in our method we use these very same dialogue evaluation metrics to learn reward $R(s_t,a_t,g)$. 
To bridge this gap, we performed the following human-in-the-loop (HITL) experiment. We first trained a pair CASPI(MINTL) models with different seeds, on 5\% of Multiwoz2.0 dataset. We then used these pair of models to predict on 0.5\% of Multiwoz2.0 train data (40 dialogues) and had a human score these pairs of generated response relative to each other. We then trained for reward $R(s_t,a_t,g)$ using pairwise causal reward learning as described in Sec:\ref{pairwiseMethod}, where examples of the mini batch are randomly sampled either from human scored examples or the ones scored by the automatic evaluation metric as show in Fig:\ref{fig:HITLLearning}. We then trained a fresh CASPI(MINTL) model on the original 5\% of data and the learnt $R(s_t,a_t,g)$. We perform human evaluation of the trained model on 24 dialogues form the test using 3 participants. Fig:\ref{fig:HITLHumanEvaluation} shows the performance. 

Though CASPI(MINTL) using just 5\% of the data outperforms DAMD trained on 100\% of data in 2 out of the 3 automatic evaluation metrics shown in Table:\ref{end2endResults} and \ref{sampleEff1}, performs poorly in human appropriateness score. With the HITL score in the reward learning, we see a boost in performance in both the human evaluation criteria: appropriateness and fluency. The 5\% data CASPI(MINTL)'s human approriateness score is now comparable to 100\% data DAMD. This goes to show the versatility of the pairwise causal reward learning. With enough richness of the neural network used, the pairwise causal reward learning can generalize to unknown dialogue evaluation criteria.

\subsection{Analysis}
\begin{figure}
    \scalebox{.27}{
    \centering
    \includegraphics{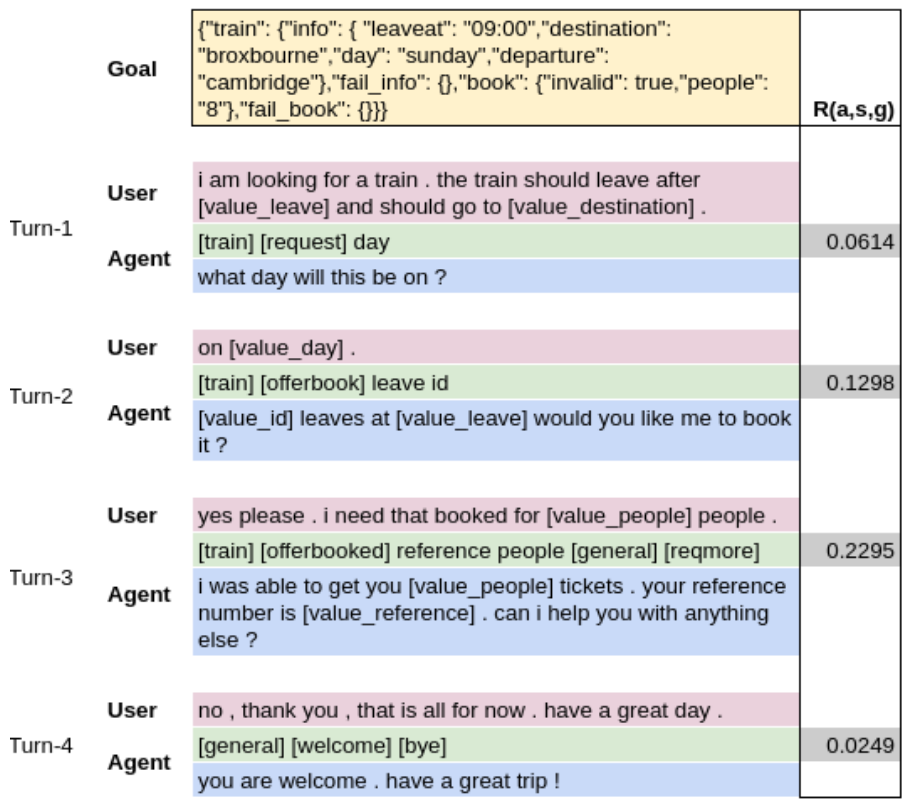}}
    \caption{Example of reward learning process}
    \label{fig:learntreward}
\end{figure} 

\begin{figure}
\centering
    \scalebox{.45}{
`    \includegraphics[width=\textwidth]{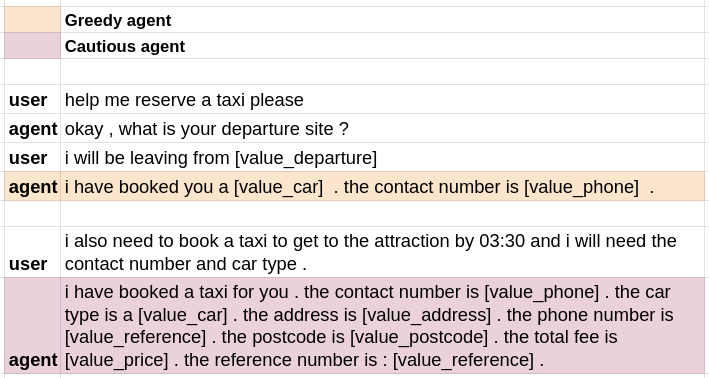}}
    \caption{Example of agent behaviour in low sample regime.}
    \label{fig:agentType}
\end{figure}%
\subsubsection{Rewards}
In this section we qualitatively analyze the results of pairwise causal reward learning. Fig:\ref{fig:learntreward} is the same conversation between a tourist and information center agents that we introduced earlier, now we have reward $R(s_t,a_t,g)$, that pairwise causal reward learning has predicted against each turn. We observe that Turn\#3 has received the highest reward, retrospectively we realize that this is the turn the transaction happens which is crucial and risk averse turn in a dialogue, which is captured by the success rate of the automatic evaluation metric. Turn\#2 gets the next best reward which captures crucial information need for transaction to happen in Turn\#3. Turn\#4 gets reward an order lower than  Turn\#3 \& 2 because other than nicety, it doesn't contribute much to the success of the conversation. It should be noted that it is typical Turn\#4 will appear in almost all conversation and in supervised learning, it'll be receiving the highest share of gradient. The learnt reward redistributes the gradient budget that is aligned to the success of the dialogue objective.

\subsubsection{Type of agents}
In this section we analyze the type of behaviour CASPI agents sometime exhibit, especially when trained in low sample regime.  

Greedy agent: In certain domains, the agents has a tendency to book a service before it has gathered all the required information or before the user requested or agreed for booking a service. The first example in Fig:\ref{fig:agentType} demonstrate this behaviour. Here the user has requested for a taxi, before enough information such as destination or time of departure are gathered, the agent books the taxi. This happens because there are gaps in automatic evaluation metrics. A low BLEU score and relatively high inform and success rate might indicate greedy agent behaviour. Other reasons for low BLEU score includes: lack of diversity in the responses or malformation of response.

Cautious agent: The agent tends to be cautious by providing long winded replies packed with more information than needed. Agent tend to do this so as not to run the risk of loosing rewards through information rate. This behaviour is demonstrated in the second example in Fig:\ref{fig:agentType}

These subtle behaviour demonstrates gap in automatic evaluation metrics. These could be weeded out using Human in the loop as described in Sec:\ref{hitlSec}.

\section{Thoughts for future work}
Appropriate choice of a metric to evaluate a rollout is crucial for learning the intention of the user. A poor choice of metrics may lead to inherited bias and the possibility of reward hacking by the policy. An option to mitigate this would be to use humans or a hybrid of metric and humans to choose between a pair of rollouts. On the flip side the use of humans might be expensive and in some cases defeat the optimization for sample complexity this work strived for. We leave this thought for future works to ponder.

\section{Conclusion}
In this work we introduced a fine grained reward learning process using an under-specified metrics function and expert demonstrations for efficiently learning Task oriented dialogue. We demonstrated the efficacy of our method on MultiWoz2.0 dataset by out performing existing state of the art method with only 10\% of data. We believe the methods is generic and can be extend to other NLP tasks.


\bibliography{example_paper}
\bibliographystyle{icml2021}



\end{document}